%% file: main.tex
\pdfoutput=1

\documentclass[11pt]{article}

\usepackage{ACL2023}

\usepackage{times}
\usepackage{latexsym}
\usepackage{graphicx}
\usepackage{booktabs}
\usepackage{array}
\usepackage{stfloats} 
\usepackage{multirow}
\usepackage{adjustbox}
\usepackage{soul}
\usepackage{color}
\usepackage{xspace}
\usepackage{multirow}
\usepackage{pifont}
\usepackage{xcolor}
\newcommand{\xmark}{\ding{55}}
\newcommand{\cmark}{\ding{51}}
\newcommand{\clic}{\textsc{CLIC}\xspace}
\newcommand*{\myfont}{\fontfamily{pcr}\selectfont}
\usepackage[T1]{fontenc}
\definecolor{palegreen}{rgb}{0.812, 0.961, 0.843}
\definecolor{paleblue}{rgb}{0.584, 0.816, 0.988}

\usepackage[utf8]{inputenc}

\usepackage{microtype}

\title{What Makes You CLIC: Detection of Croatian Clickbait Headlines}

\author{Marija Anđelić \quad Dominik Šipek \quad Laura Majer \quad Jan Šnajder \\ University of Zagreb, Faculty of Electrical Engineering and Computing  \\ TakeLab \\ \texttt{\{marija.andjelic, dominik.sipek, laura.majer, jan.snajder\}@fer.hr}}

\begin{document}
\maketitle
\begin{abstract}
Online news outlets operate predominantly on an advertising-based revenue model, compelling journalists to create headlines that are often scandalous, intriguing, and provocative -- commonly referred to as \emph{clickbait}. Automatic detection of clickbait headlines is essential for preserving information quality and reader trust in digital media and requires both contextual understanding and world knowledge. For this task, particularly in less-resourced languages, it remains unclear whether fine-tuned methods or in-context learning (ICL) yield better results. In this paper, we compile \clic, a novel dataset for clickbait detection of Croatian news headlines spanning a 20-year period and encompassing mainstream and fringe outlets. We fine-tune the BERTić model on this task and compare its performance to LLM-based ICL methods with prompts both in Croatian and English. Finally, we analyze the linguistic properties of clickbait. We find that nearly half of the analyzed headlines contain clickbait, and that finetuned models deliver better results than general LLMs.



\end{abstract}

\input{1_introduction} 
\input{2_RW}

\input{3_dataset}
\input{4_modeling}

\input{5_analysis}

\input{6_conclusion}
\input{7_limitations}

\bibliography{anthology,custom}
\bibliographystyle{acl_natbib}

\appendix
\include{appendix}

\end{document}

%% file: 1_introduction.tex
\section{Introduction}

Attention-grabbing headlines, a tactic dating back to the printed press, help publishers stand out from the competition \citep{yellowpress}. ``Clickbait'', a term defined in 2006, describes content deliberately designed to entice clicks \citep{grammarist2023clickbait} in the online landscape. Despite driving traffic, clickbait generates predominantly negative audience perception, fostering distrust toward publishers employing such tactics \citep{BLOM201587}, its manipulative nature even linking clickbait with fake news \citep{karadzhov-etal-2017-built}. It exhibits a complex relationship with sentiment \citep{chakraborty2017tabloidserasocialmedia}, characterized by hyperbolic positive terminology suggesting strategic emotional manipulation rather than genuine communication.
Driven by the negative perception of clickbait, clickbait detection -- the task of automated detection of misleading or sensationalized headlines designed to attract attention -- garnered considerable interest within the NLP community. The task progressed from feature-based linguistic approaches \citep{potthast2016} to neural architectures \citep{agrawaldeeplearning}, with transformer-based models demonstrating substantial performance improvements \citep{Zhu2023ClickbaitDV}. While primarily conducted in English, research exists for less-resourced languages including Italian  \citep{russo-etal-2024-click}, Hungarian \citep{vincze-szabo-2020-automatic}, Romanian \citep{ginga-uban-2024-scitechbaitro}, and Bulgarian \citep{karadzhov-etal-2017-built}.

The widespread use of Large Language Models (LLMs), especially using in-context learning (ICL), makes these models a reasonable candidate for clickbait detection. However, LLMs demonstrate notable performance gaps for less-resourced languages across multiple tasks \citep{rigouts-terryn-de-lhoneux-2024-exploratory, li-quantifying}, where considerably smaller Transformer models pre-trained on a specific language might outperform them \citep{ljubesic-lauc-2021-bertic}.

In this paper, we address the task of clickbait detection in the Croatian language. We introduce CLIC (\textbf{C}lickbait \textbf{L}anguage \textbf{I}dentification in \textbf{C}roatian), a novel human-annotated dataset for the task of clickbait detection. We then train a range of standard ML classifiers and Transformer-based models, and compare their performance with zero- and few-shot LLMs on this task. Additionally, we analyze linguistic features and model failures to provide deeper insight into the clickbait phenomenon.

Our work contributes a valuable new resource for a South Slavic language and deepens understanding of clickbait characteristics in Croatian media. By offering both practical detection methods and novel resources, this work makes way for clickbait neutralization in the Croatian online landscape.

%% file: 2_RW.tex
\section{Related Work}

Clickbait detection datasets are available in different languages, with English being the most represented \citep{DBLP:journals/corr/ChakrabortyPKG16, potthast-etal-2018-crowdsourcing, froebe:2023d}. Less-resourced languages are also represented. \citet{russo-etal-2024-click} constructed an Italian corpus of articles from websites known for sensationalist reporting, \citet{vincze-szabo-2020-automatic} created a small corpus downloaded from the Hungarian regional news portals, whereas \citet{ginga-uban-2024-scitechbaitro} created an annotated corpus of 10867 articles from the scientific and tech websites published on the Romanian
web. To the best of our knowledge, the only existing datasets for Slavic languages are the datasets for Bulgarian  \citep{karadzhov-etal-2017-built} and Russian \citep{APRESJAN202291}.

Prior work has also identified linguistic patterns linked to clickbait. \citet{DBLP:journals/corr/ChakrabortyPKG16} found that non-clickbait headlines are generally shorter than clickbait headlines, while \citet{biyani-secretsforclicks} found that clickbait headlines more often contain uppercase letters, question marks, quotes, exclamations, and other unusual writing patterns, suggesting they are intentionally crafted to appear more attention-grabbing.

Traditional ML methods have proven effective in clickbait detection tasks \citep{bronakowski-automatic-detection,ginga-uban-2024-scitechbaitro, DBLP:journals/corr/ChakrabortyPKG16, froebe:2023d}, demonstrating how clickbait relies on clear linguistic features.
Fine-tuned Transformer models, however, achieve superior performance, reaching F1 scores of up to 0.89 \citep{ginga-uban-2024-scitechbaitro, froebe:2023d,indurthi-etal-2020-predicting}. 

LLMs have demonstrated remarkable capabilities across NLP tasks, including clickbait detection as shown by \citet{Zhu2023ClickbaitDV}, who achieved state-of-the-art results using zero-shot and few-shot approaches. However, to our knowledge, none of those techniques have been applied to Croatian or other South Slavic languages.

%% file: 3_dataset.tex
\section{Dataset}
Since no suitable resource existed, we created a dataset of Croatian news headlines using TakeLab Retriever \citep{dukic2024takelabretrieveraidrivensearch}, an AI-driven search engine and database for Croatian news outlets. This tool covers both mainstream and fringe outlets, ensuring a diverse representation. We used the TakeLab retriever to sample data from 2000 to 2024, and unlike previous studies that specifically targeted clickbait-heavy sources, our approach sampled broadly across the entire Croatian web -- covering 32 outlets. A total of 5000 headlines were collected by extracting 200 headlines published on 25 randomly selected dates.

Eight volunteers carried out the annotation across multiple rounds, with each headline reviewed by five annotators to avoid ties. To better reflect the real-world scenario, where news consumers are often influenced by clickbait titles alone, annotators evaluated headlines without access to the full article. 
This approach differs from previous studies \citep{DBLP:journals/corr/ChakrabortyPKG16, ginga-uban-2024-scitechbaitro,vincze-szabo-2020-automatic}, which provided the annotators with full article content with the goal of clickbait neutralization. We employed a binary classification scheme with two labels (\textit{clickbait} and \textit{not clickbait}), along with an additional \textit{invalid} label to allow annotators to filter out artifacts captured during web scraping, such as navigational items, advertising content and metadata. At the end of the annotation process, due to time constraints and volunteer availability, only 3000 randomly selected headlines were selected from a total of 30 news portals out of the original 5000. The resulting annotated dataset is made publicly available in its entirety.\footnote{Dataset available at: \url{https://takelab.fer.hr/data/clic}} 

Out of the 3,000 total annotated headlines (examples shown in Table \ref{tab:LLMerrors}), 77 were labeled as \textit{invalid}, and 16 were duplicates. This resulted in a final dataset of 2,907 annotated headlines, with 1,536 labeled \textit{clickbait} ($ 52.84\%$) and 1,371 ($ 47.16\%$) as \textit{not clickbait}, making the corpora relatively balanced and in accordance with previous work \citet{ginga-uban-2024-scitechbaitro, DBLP:journals/corr/ChakrabortyPKG16}. The annotators achieved an inter-annotator agreement of 0.53 using the Fleiss-kappa score, categorized as moderate agreement. This confirms the subjective nature of clickbait detection, both in its linguistic ambiguity and in how readers perceive manipulative techniques.

Observing the clickbait distribution over the years, the clickbait to non-clickbait headline ratio is consistent at around $50\%$ (Fig.~\ref{fig:clickbait-over-years}), indicating no significant increase in clickbait on our sample.

%% file: 4_modeling.tex
\section{Clickbait Detection}

\paragraph{Baselines.}
As baselines, we use several standard ML models, all utilizing TF-IDF vectorization of headlines as input features: logistic regression, SVM with a linear kernel (both optimized via GridSearchCV), and a simple neural network with two hidden layers using ReLU activation and sigmoid output layer for binary classification. For all experiments, we use an 80-10-10 split for training, validation, and testing, respectively. We also report the majority class baseline (0.55), representing the accuracy achievable by predicting the most frequent class in our test set.

\paragraph{Fine-tuned Transformers.}
We finetune BERT \citep{devlin2019bertpretrainingdeepbidirectional} and BERTić \citep{ljubesic-lauc-2021-bertic}, an Electra-based Transformer model pre-trained on south-Slavic languages, including Croatian.

\paragraph{LLMs.}
For our experiments, we use ICL with both zero-shot and few-shot prompting for various LLMs, including the closed-source GPT 4.1, and smaller open-source models Gemma7B, Phi-3.5-mini, and Mistral7B-Instruct. 

For prompting the models, we use various configurations based on the amount of instructions and demonstrations added to the naive prompt (containing just the instruction to classify clickbait). To do so, we draft prompt components: clickbait definition (D), features of clickbait articles (F), and few-shot examples (E) containing both clear and ambiguous cases. We construct combinations of the D, F, and E components to be able to isolate which component attributes mostly to model performance. Also, we translate the prompts to Croatian to compare whether the model performance is higher using English prompts or using the same language as the examples are in. The full prompts are available in Appendix~\ref{tab:CB}.

\begin{table}
    \centering
    \small
    \begin{tabular}{lcc}
        \toprule
        \textbf{Model} & \textbf{Accuracy} & \textbf{F1} \\
        \midrule
        Logistic regression       & 0.63 & 0.67 \\
        Simple neural network     & 0.59 & 0.61 \\
        SVM                       & 0.63 & 0.66 \\
        BERT-base-multilingual      & 0.72 & 0.71 \\
        BERTić                    & \sethlcolor{palegreen}\hl{0.78} & \sethlcolor{palegreen}\hl{0.78} \\
        \bottomrule
    \end{tabular}
    \caption{Performance comparison of various models}
    \label{tab:model-results}
\end{table}

\paragraph{Results.}
Table~\ref{tab:model-results} shows the classification performance of all implemented models. All tested ML methods outperform both the majority and random baselines, confirming that clickbait detection fundamentally relies on linguistic patterns. While our baseline models rely solely on TF-IDF vectorization for feature extraction, without the sophisticated linguistic feature engineering employed in prior work (e.g., \citet{potthast2016} and \citet{chakraborty2017tabloidserasocialmedia} used structural, lexical, and syntactic features), we still achieve reasonable performance. This suggests that even simple lexical features can capture many of the distinguishing patterns of clickbait in Croatian.

The fine-tuned Transformer models achieved F1 scores of $71.4\%$ (BERT) and $77.6\%$ (BERTić) respectively, with BERTić expectedly coming out on top. These results coincide with the findings of previous authors.

The results for LLMs with English prompts are shown in Table~\ref{tab:LLM-f1}. All tested LLMs outperform both the majority and random baselines, and perform comparably to the ML baseline. However, compared to the results of a fine-tuned BERTić model, all of the LLMs underperform. Gemma7B is the most consistent out of the group with regards to prompt design and incorporation of clickbait features. In all models, the inclusion of clickbait features has been shown to have the most drastic effect on the model's eventual performance, and combining features with few-shot examples produces the highest performance. The importance of examples can be seen when examining the poor performance of a combination of only definition and features (DF), which underscores the critical role of few-shot learning approaches in this task. All of the models show a significant discrepancy between results for prompts in Croatian and English. An analysis of precision versus recall reveals a clear difference between languages. When prompted in English, models tend to be more liberal, favoring recall, whereas prompts in other languages lead models to prioritize precision.

\begin{table}[t]
\centering
\scriptsize
\setlength{\tabcolsep}{3pt} 
\begin{tabular}{@{}lcccccccccc@{}}
\toprule
\textbf{} & \textbf{Model} & \textbf{Naive} & \textbf{D} & \textbf{E} & \textbf{F} & \textbf{DE} & \textbf{DF} & \textbf{FE} & \textbf{DFE} \\
\midrule
\textit{English} & Mistral 7B  & 0.66 & 0.63 & 0.69 & 0.43 & 0.71 & 0.46  & \sethlcolor{palegreen}\hl{0.73} & 0.72 \\
& Phi 3.5     & \sethlcolor{palegreen}\hl{0.70} & 0.60 & 0.56 & 0.52 & 0.59 & 0.45 & 0.62 & 0.57 \\
& gemma 7b    & 0.70 & 0.70 & 0.67 & 0.70 & 0.69 & 0.70 & \sethlcolor{palegreen}\hl{0.71} & 0.69 \\
& gpt-4.1     & 0.60 & 0.43 & \sethlcolor{palegreen}\hl{0.68} & 0.36 & 0.46 & 0.30 & 0.50 & 0.41 \\
\midrule
\textit{Croatian} & Mistral 7B  & 0.42 & \sethlcolor{palegreen}\hl{0.64} & 0.25 & 0.61 & 0.53 & 0.58 & 0.60 & 0.55 \\
& Phi 3.5     & 0.44 & 0.49 & 0.44 & \sethlcolor{palegreen}\hl{0.66} & 0.48 & 0.63 & 0.59 & 0.59 \\
& gemma 7b    & \sethlcolor{palegreen}\hl{0.70} & 0.50 & 0.62 & 0.67 & 0.59 & 0.57 & 0.62 & 0.67 \\
& gpt-4.1     & \sethlcolor{palegreen}\hl{0.63} & 0.54 & 0.59 & 0.34 & 0.49 & 0.41 & 0.50 & 0.45 \\
\bottomrule
\end{tabular}
\caption{F1 scores for various configurations (best in green); D = definition, F = features, E = explanation}
\label{tab:LLM-f1}
\end{table}

%% file: 5_analysis.tex
\section{Analysis}

\paragraph{Feature Analysis.}
We conducted computational and linguistic analyses to identify the linguistic patterns that distinguish clickbait from legitimate headlines and understand our models' detection mechanisms. Initial attention masking and saliency analysis failed to yield coherent patterns shown in Table~\ref{tab:attention-weights}, with random words receiving high importance scores. Therefore, we employed Part of Speech (PoS) tagging, revealing clear statistical differences between clickbait and non-clickbait content. Results in Table \ref{tab:pos-tags} show pronouns, auxiliary words, and determiners are significantly overrepresented in clickbait articles (p<0.05, chi-squared tests), aligning with findings by \citet{ginga-uban-2024-scitechbaitro}. Interjections appear exclusively in clickbait. Chi-squared tests confirm statistical significance for most differences, with determiners, particles, and pronouns showing the strongest clickbait association. Conversely, non-clickbait articles contain significantly more proper nouns and numbers.

\begin{table}[t]
    \centering
    \scriptsize
    \begin{tabular}{p{2.4cm}c|p{2.4cm}c}
        \toprule
        \multicolumn{2}{c|}{\textbf{Clickbait headlines}} & \multicolumn{2}{c}{\textbf{Non-clickbait headlines}} \\
        \midrule
        \textbf{Part of speech} & \textbf{Ratio} & \textbf{Part of speech} & \textbf{Ratio} \\
        \hline
        Interjection & $\infty$ & Proper noun & 1.54$\times$ \\
        Determiner & 2.32$\times$ & Symbol & 1.44$\times$ \\
        Particle & 2.14$\times$ & Number & 1.31$\times$ \\
        Pronoun & 1.90$\times$ & Other & 1.29$\times$ \\
        Auxiliary verb & 1.53$\times$ & Adposition & 1.19$\times$ \\
        \bottomrule
    \end{tabular}
    \caption{Relative frequency of part-of-speech tags in clickbait vs. non-clickbait headlines. Values represent the frequency with which each tag appears in its respective category.}
    \label{tab:pos-tags}
\end{table}

\begin{figure}[t]
    \centering
    \includegraphics[width=\linewidth]{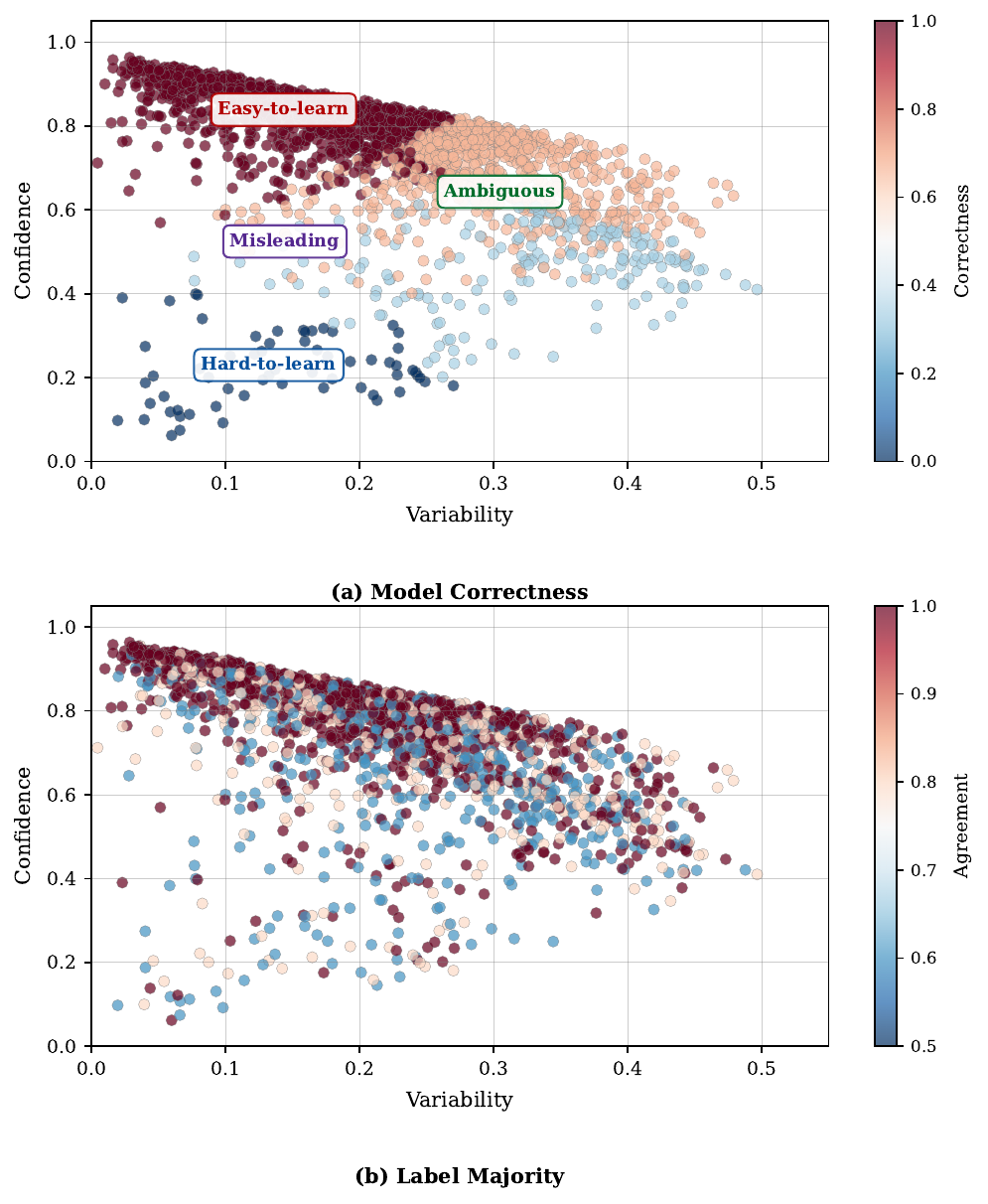}
    \caption{Dataset cartography visualization showing the classification of headlines by learnability. (a) Model correctness indicates how well the model performs on various examples. (b) Label majority displays the level of agreement among annotations.}
    \label{fig:data-cartography}
\end{figure}

\paragraph{Dataset Cartography.} 
The dataset cartography method, proposed by \citet{swayamdipta-etal-2020-dataset}, enables the identification of hard-to-learn instances during training, as well as ambiguous and potentially mislabeled instances. Figure~\ref{fig:data-cartography} shows the dataset cartography for the fine-tuned BERTić model.
By analyzing confidence, variability, and correctness across training epochs, we gain insight into training dynamics, and cluster the train-set instances into regions -- \emph{easy-to-learn}, \emph{ambiguous}, \emph{misleading}, and \emph{hard-to-learn}. 

In cases where variability is high and confidence is around 0.5, we identify the space of \textit{ambiguous} instances. Upon inspection, those instances are characterized by balanced clickbait elements, contextual dependency, and mixed use of language devices. The \textit{misleading} region represents instances with low variability and medium confidence. These instances mostly lack classic clickbait markers and are mostly short, direct statements.

\paragraph{Comparison with Annotators.}
Since clickbait is a subjective classification task, where instances with lower agreement could indicate complex or ambiguous instances, we compare BERTić and LLM performance with annotator agreement levels. For BERTić, we again utilize dataset cartography, but instead of the \textit{correctness} dimension, we display the majority proportion for the given instance. Figure~\ref{fig:data-cartography}(b) shows no distinct regions. This suggests a decoupling between human label variation and model difficulty, meaning instances that humans find ambiguous are not necessarily the same ones that challenge the model. 

\paragraph{Error analysis.}
For a qualitative error analysis of LLMs, we construct an intersection of misclassified instances across prompt variants, then analyze the resulting subset. Examples are shown in Table~\ref{tab:LLMerrors}. For non-clickbait instances consistently labeled as clickbait, we find that they mostly consist of sensationalist and dramatic expressions used in factual headlines -- exclamation marks, quotation marks, quotes, all caps, numbers -- which are features mostly appearing in clickbait instances. For instances that annotators labeled clickbait, but consistently classified as non-clickbait, we find no sensational or shocking expressions, the tone is neutral and serious, but the full context is missing. Concealing crucial information is a clear characteristic of clickbait articles, leading to these errors. 

\begin{table}[t]
\centering
\scriptsize
\setlength{\tabcolsep}{3pt}
\begin{tabular}{@{}p{5.8cm}cc@{}}
\toprule
\textbf{Title} & \textbf{True} & \textbf{Pred} \\
\midrule
Senzacija: Niko Kovač novi trener Bayerna & \xmark & \cmark \\
Pametna kuna bira najbolju poslovnu ideju! Prijavite se! & \xmark & \cmark \\
Irak: U 24 sata ubijena petorica američkih vojnika & \xmark & \cmark \\
\midrule
Lalovac ne pada daleko od Linića & \cmark & \xmark \\
Urednik mu nije trebao. Znao je sve o pisanju & \cmark & \xmark \\
Sin na listi HDZ-a, a otac u izbornom povjerenstvu & \cmark & \xmark \\
\bottomrule
\end{tabular}
\caption{Examples of misclassified headlines. \cmark = clickbait, \xmark = not clickbait}
\label{tab:LLMerrors}
\end{table}

%% file: 6_conclusion.tex
\section{Conclusion}
In this paper, we addressed the task of clickbait detection for Croatian with a new dataset. We evaluated LLM-based ICL methods and fine-tuned BERTić against traditional ML approaches. Our experiments show that fine-tuned BERTić achieves the best overall performance, while LLMs improve when clickbait features are explicitly included in prompts.

%% file: 7_limitations.tex
\section{Limitations and Risks}
\paragraph{Limitations.}
Our current approach focuses solely on headline annotation for clickbait detection. This method, while effective for initial classification, does not allow for an assessment of the semantic relationship between headlines and their corresponding article content. Consequently, headlines that are sensationalized but ultimately accurate might be misclassified. Additionally, our findings are based on Croatian-language data. Therefore, the generalizability of these results to other languages or cultural contexts, which may have distinct clickbait conventions, remains to be explored.

\paragraph{Risks.}
The linguistic patterns identified in this work could enable more sophisticated clickbait generation that evades detection systems, potentially exacerbating rather than mitigating the problem. Another possible risk is that, in the event of deploying our models, there is a risk of misclassifying legitimate news as clickbait, potentially suppressing real journalism. If any biases are present, they may be amplified in the automatic content filtering process.

%% file: appendix.tex
\begin{table*}
    \centering
    \begin{adjustbox}{width=\textwidth}
    {\small
    \begin{tabular}{c c} 
     \toprule
     \textbf{Level} & \textbf{Prompt} \\ 
     \midrule
    English & \parbox{\textwidth}{
  \myfont{
     \hl{Clickbait headlines involve subtle and manipulative techniques to attract attention and pique readers' curiosity so they click on the article. It is important to distinguish clickbait from irrelevance, where uninteresting or gossipy headlines are not necessarily clickbait.}
    \\ \sethlcolor{pink}\hl{Clickbait headlines often contain the following features:
        1) Sensationalism - Does it use emotional words like "shocking," "unbelievable," or "must-see"?
        2) Missing Information - Does it leave out key details, forcing the reader to click?
        3) Manipulative Language - Does it promise "one simple trick" or "secrets they don’t want you to know"?}
     \\Is the following headline clickbait? Answer with Yes or No.
     {
    \\ \sethlcolor{paleblue}\hl{Headline: "Što se događa sa slavnom pjevačicom: Novi imidž razočarao fanove" Answer: Yes
    \\ Headline: "Policajac tužio svog načelnika za klevetu" Answer: Yes
    \\Headline: "Vaterpolisti dubrovačkog Juga osvojili Hrvatski kup" Answer: No
    \\Headline: "Sudarila se jahta s trajektom kod Biograda: 'Grunulo je, djeca su plakala, prestravili smo se...'" Answer: No
    }  {headline} Answer:} 

 }}\\
      \midrule
    Croatian & \parbox{\textwidth}{
  \myfont{
     \hl{Clickbait naslovi uključuju suptilne i manipulativne tehnike kojima se privlači pozornost i potiče znatiželja čitatelja kako bi kliknuli na članak. Važno je razlikovati clickbait s irelevantnošću pri čemu nezanimljivi ili trač naslovi ne moraju nužno biti clickbait.}
    \\ \sethlcolor{pink}\hl{Clickbait naslovi često sadržavaju sljedeće značajke:
        1) Senzacionalizam - Koristi li emocionalne riječi poput "šokantno", "nevjerojatno" ili "morate vidjeti"?
        2) Izostavljanje informacija - Izostavlja li ključne pojedinosti, tjera li čitatelja da klikne?
        3) Manipulativni jezik - Obećava li "jedan jednostavan trik" ili "tajne koje ne žele da znate"?}
     \\Je li navedeni naslov clickbait naslov? Odgovori s Da ili Ne.
     {
    \\ \sethlcolor{paleblue}\hl{Headline: "Što se događa sa slavnom pjevačicom: Novi imidž razočarao fanove" Odgovor: Da
    \\ Headline: "Policajac tužio svog načelnika za klevetu" Odgovor: da
    \\Headline: "Vaterpolisti dubrovačkog Juga osvojili Hrvatski kup" Odgovor: Ne
    \\Headline: "Sudarila se jahta s trajektom kod Biograda: 'Grunulo je, djeca su plakala, prestravili smo se...'" Odgovor: No
    }  {headline} Odgovor:} 
 }}\\
     \bottomrule
    \end{tabular}}
    \end{adjustbox}
    \captionof{table}{System prompts in English and Croatian used for inference in EFD (D = definition, F = features, E = explanation) configuration.}
\label{tab:CB}
\end{table*}

\begin{figure}[htbp]
    \centering
    \includegraphics[width=\linewidth]{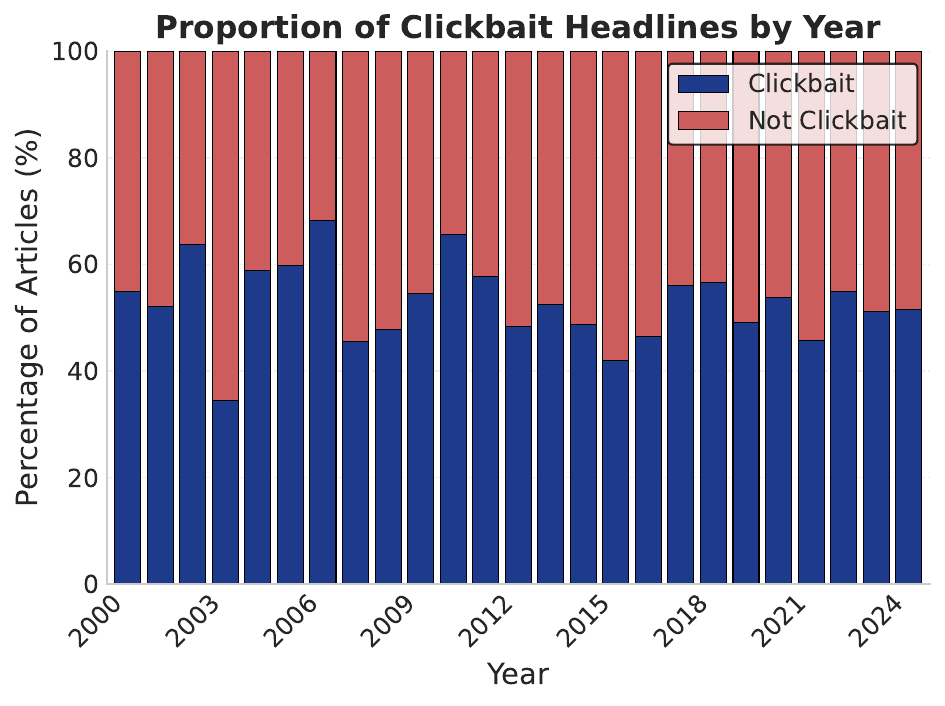}
    \caption{Percentage of news headlines classified as clickbait versus non-clickbait for each year in the dataset.}
    \label{fig:clickbait-over-years}
\end{figure}

\clearpage
\begin{table}[t]
\centering
\caption{Top 20 influential words based on attention mapping (mean IG scores)}
\begin{tabular}{lrl}
\toprule
\textbf{Croatian Word} & \textbf{Mean IG} & \textbf{English Translation} \\
\midrule
cvatu & 395.6445 & bloom/flourish \\
Glumica & 14.8237 & actress \\
loš & 12.4242 & bad \\
zajedničke & 11.3327 & joint/common \\
ekonomija & 10.7566 & economy \\
Ruši & 10.4093 & demolishes/topples \\
njega & 10.3574 & him/his \\
zločinačka & 8.7656 & criminal \\
nezainteresirani & 7.0193 & uninterested \\
FUUUUUUJ & 6.6199 & expression of disgust \\
Pokrenut & 6.5441 & launched/initiated \\
petka & 6.3973 & Friday \\
SAČIĆ & 6.0077 & surname \\
Paltrow & 5.8773 & surname (Paltrow) \\
vječna & 5.8190 & eternal \\
oko & 5.5182 & eye/around \\
zadnje & 5.4024 & last/final \\
autogol & 5.2721 & own goal \\
stampeda & 5.2050 & stampede \\
Satelit & 4.8463 & satellite \\
\bottomrule
\end{tabular}
\label{tab:attention-weights}
\end{table}

\begin{table}[t]
    \centering

    \begin{tabular}{lc}
        \toprule
        \textbf{News Outlet} & \textbf{Number of Articles} \\
        \midrule
        vecernji.hr & 505 \\
        glas-slavonije.hr & 394 \\
        index.hr & 346 \\
        24sata.hr & 330 \\
        net.hr & 278 \\
        dnevnik.hr & 209 \\
        jutarnji.hr & 182 \\
        tportal.hr & 155 \\
        slobodnadalmatija.hr & 103 \\
        narod.hr & 76 \\
        hrt.hr & 58 \\
        direktno.hr & 55 \\
        hr.n1info.com & 38 \\
        dnevno.hr & 35 \\
        novilist.hr & 30 \\
        rtl.hr & 25 \\
        lupiga.com & 24 \\
        h-alter.org & 20 \\
        telegram.hr & 16 \\
        geopolitika.news & 6 \\
        \bottomrule
    \end{tabular}
    \caption{Number of articles by news outlet (minimum 5 articles per outlet).}
    \label{tab:articles-by-outlet}
\end{table}